\documentclass[letterpaper,twocolumn,10pt, table,xcdraw]{article}
\usepackage{style/usenix2019_v3}
\usepackage{style/mist}

\usepackage{lipsum}

\begin{document}

\title{\Large \bf A Survey on Simulators for Testing Self-Driving Cars}
\author{Prabhjot Kaur}
\author{Samira Taghavi}
\author{Zhaofeng Tian}
\author{Weisong Shi}

\affil{Department of Computer Science, Wayne State University, Detroit, USA}  
\date{\today}

\maketitle

\begin{abstract}
A rigorous and comprehensive testing plays a key role in training self-driving cars to handle variety of situations that they are expected to see on public roads. 

The physical testing on public roads is unsafe, costly, and not always reproducible. This is where testing in simulation helps fill the gap, however, the problem with simulation testing is that it is only as good as the simulator used for testing and how representative the simulated scenarios are of the real environment. In this paper, we identify key requirements that a good simulator must have. Further, we provide a comparison of commonly used simulators. Our analysis shows that CARLA and LGSVL simulators are the current state-of-the-art simulators for end to end testing of self-driving cars for the reasons mentioned in this paper. Finally, we also present current challenges that simulation testing continues to face as we march towards building fully autonomous cars.
\end{abstract}
\section{Introduction}
\label{sec:intro}

According to the annual Autonomous Mileage Report $\footnote{\href{http://bit.ly/39tf0QT}{Autonomous Mileage Report: Disengagement Reports}}$ published by the California Department of Motor Vehicles, Waymo has logged billions of miles in testing so far. As of 2019, the company’s self-driving cars have driven 20 million miles on public roads in 25 cities and additionally 15 billion miles through computer simulations \cite{waymosimulation:00}. The number of miles driven is important, however,  it is the sophistication and diversity of miles accumulated that determines and shapes the maturity of the product \cite{7795548}. While the real world testing through physical driving tests is not replaceable without the extra cost required to build the infrastructure and in some cases jeopardizing the safety of public \cite{uber:00}, simulation plays a key role in supplementing and accelerating the real world testing\cite{waymosimulation:00}. It allows one to test scenarios that are otherwise highly regulated on public roads because of safety concerns \cite{9046805}. It is reproducible, scalable and cuts the development. 

There are many simulators available for testing the software for self-driving cars, which have their own pros and cons. Some of them include CarCraft and SurfelGAN used by Waymo \cite{waymocameradata}, \cite{fadaie2019state}, Webviz and The Matrix used by Cruise, and DataViz used by Uber \cite{autosimulation}. Most of these are proprietary tools, however, there are many open source simulators available as well. In this paper, we compare MATLAB/Simulink, CarSim, PreScan, Gazebo, CARLA and LGSVL simulators with the objective of studying their performance in testing new functionalities such as perception, localization, vehicle control, and creation of dynamic 3D virtual environments. Our contribution is two-fold. We first identify a set of requirements that an ideal simulator for testing self-driving cars must have. Secondly, we compare the simulators mentioned above and make the following observations.

An ideal simulator is the one that is as close to reality as possible. However, this means it must be highly detailed in terms of 3D virtual environment and very precise with lower level vehicle calculations such as the physics of the car. So, we must find a trade off between the realism of the 3D scene and the simplification of the vehicular dynamics \cite{figueiredo2009approach}. CARLA and LGSVL meet this trade-off, making them the state-of-the-art simulators. Further, Gazebo is also a popular robotic 3D simulator but it is not very efficient in terms of time involved in creating a 3D scene in the simualtion environment. The simulators such as MATLAB/Simulink still play a key role because they offer detailed analysis of the results with their plotting tools. Similarly, CarSim is highly specialized at vehicle dynamic simulations as it is backed by more precise car models. The detail reasoning behind these observations in described in the paper below.

This paper is organized into several sections. Section 2 provides a summary of evolution of automotive simulators followed by Section 3 which identifies and describes requirements for an automotive simulator used for testing self-driving cars. Then, it provides a survey of several open source simulators in Section 4, followed by comparison of these simulators in Section 5. Section 6 provides various challenges that the simulators currently have. Finally this paper concludes in Section 8.

\section{Motivation and Background}
\label{sec:motivation}

The complexity of automotive software and hardware is continuing to grow as we progress towards building self-driving cars. In addition to tradition testing such as proper vehicle dynamics, crash-worthiness, reliability, and functional safety, there is a need to test self-driving related algorithms and software, such as deep learning and energy efficiency \cite{liu2020computing}. As an example, a Volvo vehicle built in 2020 has about 100 million lines of code according to their data \cite{antinyan2020revealing}. This includes code for transmission control, cruise control, collision mitigation, connectivity, engine control and many other basic and advanced functionalities that come with the cars bought today. Similarly, the cars now have more advanced hardware, which includes plethora of sensors that ensure vehicles are able to perceive the world around them just like humans do\cite{ hirz2018sensor}. Therefore, the complexity of the modern age vehicle is the result of both more advanced hardware and software needed to process the information retrieved from the environment and for decision making capability.

Finally, in order to assure that the finished product complies with the design requirements, it must pass rigorous testing which is composed of many layers. It ranges from lower level testing of Integrated Circuits (ICs) to higher level testing of vehicle behavior in general. The testing is accomplished by relying on both physical and simulation testing. As the complexity and functionality of vehicles continue to grow, so does the complexity, scale and the scope of testing that becomes necessary. Therefore, the simulators used for automotive testing are in a continuous state of evolving.  

These simulators have evolved from merely simulating vehicle dynamics to also simulating more complex functionalities. Table~\ref{tab:simevolution} shows various levels of automation per the Society of Automotive Engineers (SAE) definitions \cite{standardj3016}, along with the evolving list of requirements for testing that are inherent in our path to full automation. It is important to note that Table~\ref{tab:simevolution} focuses on requirements that are essentially new to testing driver assisted features and autonomous behavior \cite{schoner2017role}. This includes things such as perception, localization and mapping, control algorithms and path planning 
\begin{table}[h!]
\centering
\caption{Testing Requirements to meet SAE Automation Levels.}
\label{tab:simevolution}
\scalebox{0.77}{
\begin{threeparttable}[b]
\begin{tabular}{|c|c|c|c|}
\hline
\multicolumn{2}{|c|} {\textbf{SAE J3016  Levels of Driving Automation}} & \textbf{Testing Requirements}\\ \hline
\textbf{Levels} & \textbf{Description} &  \\ \hline
Level 0 & \begin{tabular}[l]{@{}c@{}}No Automation: Features are limited to,\\ warnings and momentary assistance.\\ Examples: LDW, Blind Spot Warning \end{tabular}  & \begin{tabular}[l]{@{}c@{}}Simulation of: Traffic flow, \\multiple road terrain type,\\ radar and camera sensors \end{tabular}\\ \hline
Level 1 &  \begin{tabular}[l]{@{}c@{}}Assisted: Features provide steering OR\\ brake/acceleration control.\\ Examples: Lane Centering OR ACC \end{tabular} & \begin{tabular}[l]{@{}c@{}}All of the above plus\\Simulation of: \\vehicle dynamics, \\ultrasonic sensors \end{tabular} \\ \hline
Level 2 &  \begin{tabular}[l]{@{}c@{}}Partial Automation: Features provide \\steering AND brake/acceleration control.\\ Examples: Lane Centering AND ACC \\ at the same time\end{tabular}& \begin{tabular}[l]{@{}c@{}}All of the above plus\\Simulation of: \\driver monitoring system.\\Human-machine interface \end{tabular} \\ \hline
Level 3 & \begin{tabular}[l]{@{}c@{}}Conditional Automation: Features can \\ drive the vehicle when all of its\\ conditions are met. \\ Examples: Traffic Jam Assist\\ \end{tabular} & \begin{tabular}[l]{@{}c@{}}All of the above plus \\Simulation of: Traffic\\ infrastructure,dynamic objects \end{tabular} \\ \hline
Level 4 & \begin{tabular}[l]{@{}c@{}}High Automation: Features can drive\\ the vehicle under limited conditions.\\ No driver intervention. \\ Examples: Local Driverless taxis\\ \end{tabular} & \begin{tabular}[l]{@{}c@{}}All of the above plus \\Simulation of: different \\ weather conditions,\\ lidar, camera, radar sensors, \\mapping and localization \end{tabular}\\ \hline
Level 5 & \begin{tabular}[l]{@{}c@{}}Full Automation: Features can drive \\the vehicle in all conditions and\\ everywhere. \\Examples: Full autonomous vehicles\\ everywhere \end{tabular} & \begin{tabular}[l]{@{}c@{}}All of the above plus\\compliance with all the road, \\ rules, V2X communication \end{tabular}\\ \hline
\end{tabular}
\begin{tablenotes}
    \item[1] These definitions are per the SAE J3016 Safety levels. More details can be found below: 
    \item[] \href{https://www.sae.org/standards/content/j3016_201806/}{https://www.sae.org/standards/content/j3016\_201806/} 
    \item[2] LDW = Lane Departure Warning
    \item[3] ACC = Adaptive Cruise Control
\end{tablenotes}
\end{threeparttable}
}
\vspace{-0.6 cm}
\end{table}

 Thus, the simulators intended to be used for testing self driving cars must have requirements that extend from simulating physical car models to various sensor models, path planning and control. Section 3 dives deeper into these requirements.

\section{Methodology}
\label{sec:methodlogy}
The emphasis of this paper is on testing the new and highly automated functionality that is unique to self-driving cars. This section identifies a set of criteria that can serve as a metric to identify which simulators are a best fit for the task at hand. 

The approach we take to compile requirements for a simulator is as below. Firstly, we focus on the requirements driven by the functional architecture of self-driving cars \cite{fan2019key} (Requirements 1-4). Secondly, we focus on the requirements that must be met in order to support the infrastructure to drive the simulated car in (Requirements 5-7). Thirdly, we define the requirements that allow the use of simulator for secondary tasks such as data collection for further use (Requirement 8). Finally, we list generic requirements desired from any good automotive simulator (Requirement 9).

\begin{enumerate}
	\item \textbf{Perception}: One of the vital components of self-driving cars is its ability to see and make sense (perceive) the world around itself. This is called perception. The vehicle perception is further composed of hardware, that is available in the form of wide variety of automotive grade sensors and software, that interprets data collected by various sensors to make it meaningful for further decisions. The sensors that are most prevalent in research and commercial self-driving cars today include camera, LiDAR, ultrasonic sensor, radar, Global Positioning System (GPS), Inertial Measurement Unit (IMU) \cite{van2018autonomous}. In order to test a perception system, the simulator must have realistic sensor models and/or be able to support an input data stream from the real sensors for further utilization. Once the data from these sensors is available within the simulation environment, researchers can then test their perception methods such as sensor fusion  \cite{kocic2018sensors}. The simulated environment can also be used to guide sensor placement in a real vehicle for optimal perception.
		
	\item \textbf{The multi-view geometry }
	The Simultaneous Localization and Mapping (SLAM) is one of  components of Autonomous Driving (AD) systems that focuses on constructing  the map of unknown environments and tracking the location of the AD system inside the updated map. In order to support SLAM applications, the simulator should provide the intrinsic and extrinsic features of cameras. In other words, it should provide the camera calibration. According to this information, the SLAM algorithm can run the multi-view geometry and estimate the camera pose and localize AD system inside the global map.
	
	\item \textbf{Path Planning}:The problem of path planning revolves around planning a path for a mobile agent so that it is able to move around autonomously without collision with its surroundings. The path planning problem for autonomous vehicles piggy backs on the research that has already been done in the field of mobile robots in the last decade. This problem is sub-divided into local and global planning \cite{ITS} where the global planner is typically generated based on a static map of the environment and the local planner is created incrementally based on the immediate surroundings of the mobile agent. In order to create these planners, various planning algorithms play a key role \cite{Gmp}. To implement such intelligent path planning algorithms like A*, D* and RRT algorithms\cite{ITS}, the simulator should at least have a built-in function to build maps or have interfaces for importing maps from outside. In addition, the simulator should have interfaces for programming customized algorithms.
	\item \textbf{Vehicle Control}: The final step after a collision free path is planned is to execute the predicted trajectory as closely as possible. This is accomplished via the control inputs such as throttle, brake and steering \cite{fan2019key} that are monitored by closed loop control algorithms \cite{distance}. The Proportional–integral–derivative (PID) control algorithm and Model Predictive Control (MPC) algorithm are commonly seen in research and industries \cite{MPC}. To implement such intelligent control algorithms, the simulator should be capable of building vehicle dynamic models and programming the algorithms in mathematical forms.

	\item \textbf{3D Virtual Environment}: In order to test various functional elements of a car mentioned in the above requirements, it is equally important to have a realistic 3D virtual environment. The perception system relies on photogenic view of the scene to sense the virtual world. This 3D virtual environment must include both static objects such as buildings, trees, etc. and dynamic objects such as other vehicles, pedestrians, animals, and bicyclists. Furthermore, the dynamic objects must behave realistically to reflect the true behavior of these dynamic entities in an environment. 
	
	In order to achieve 3D virtual environment creation, simulators can either rely on game engines or use the High Definition (HD) map of a real environment and render it in a simulation \cite{fadaie2019state}. Similarly, in order to simulate dynamic objects, the vehicle simulators can leverage other domains such as pedestrian models \cite{camara2020pedestrian} to simulate realistic pedestrians movement in the scene. Furthermore, the 3D virtual environment must support different terrains and weather conditions that are typical in a real environment.
	
	It is important to note that the level of detail in a 3D virtual environment depends on the simulation approach taken. Some companies such as Uber and Waymo do not use highly detailed simulators \cite{fadaie2019state}. Therefore, they do not use simulators to test perception models. However, if the goal is to test perception models in simulation, then the level of detail is very important.
	\item \textbf{Traffic Infrastructure}: In addition to the requirements for a 3D virtual environment mentioned above, it is also important for a simulation to have the support for various traffic aids such as traffic lights, roadway sinage, etc. \cite{lafuente2005traffic}. This is because these aids help regulate traffic for the safety of all road users. It is projected that the traffic infrastructure will evolve to support connected vehicles in the near future \cite{liu2019systematic}. However, until the connected vehicles become a reality, self-driving cars are expected to comply with the same traffic rules as the human drivers. 
	
	\item \textbf{Traffic Scenarios Simulation}: The ability to create various traffic scenarios is one of main points that identifies whether a simulator is valuable or not. This allows the researchers to not only re-create/play back a real world scenario but also allows them to test various "what-if" scenarios that cannot be tested in a real environment because of safety concerns. This criteria considers not only the variety of traffic agents but also the mechanisms that the simulator provides to generate these agents. Different types of dynamic objects consist of humans, bicycles, motorcycles, animals, and  vehicles such as buses, trucks, ambulances and motorcycles. In order to  generate scenes close to real world scenes, it is important that simulator supports significant number of these dynamic agents. In addition, simulator should provide a flexible API that allows users to manage different aspects of simulation which consists of generating traffic agents and more complex scenarios such as pedestrian behaviors, vehicles crashes, weather conditions, sensor types, stops signs, and etc. 
	
	\item \textbf{2D/3D Ground Truth}
	In order to provide the training data to the AI models, the simulator should provide object labels and bounding boxes of the objects appearing in the scene. The sensor outputs each video frame where objects are encapsulated in a box. 
	\item\textbf{ Non-functional Requirements} 
    The qualitative analysis of open source simulators includes different aspects that can help AD developers to estimate the learning time and the duration required for simulating different scenarios and experiments.   
    
    \textbf{Well maintained/Stability}
    In order to use simulator for different experiments and testing, the simulator should have comprehensive documentation that makes it easy to use. In case that maintenance teams improve the simulator, if the backward compatibility is not considered, the documentation should provide the precise mapping between the deprecated APIs and newly added APIs.
    
    \textbf{Flexibility/Modular} 
    Open source simulators should follow division of concept principle that can help AD developers to leverage and extend different scenarios in shorter time. In addition, the simulator can provide a flexible API that enables defining customized version of sensors, generating new environments and adding different agents. 
    
    \textbf{Portability}
    If the simulator is able to run on different types of operating systems, it enables users to leverage the simulator more easily. Most of users may not have access to the different types of operating systems at the same time, therefore the simulators portability can save the time for the users.  
    
    \textbf{Scalability via a server multi-client architecture}
    Scalable architecture such as client-server architecture enables multiple clients run on different nodes to control different agents at the same time. This is helpful specifically for simulating the congestion and/or complex scenes. 
    
    \textbf{Open-Source} It is preferred that a simulator be open source. The open source simulators enable more collaboration, collective progress and allows to incorporate learning from peers in the same domain.

\end{enumerate}

\section{Simulators}
This section provides a brief description of simulators that were analyzed and compared.

\subsection{MATLAB/Simulink}
MATLAB/Simulink published Automated Driving Toolbox™, which provides various tools that facilitate the design, simulation and testing of  Advanced Driver Assisted Systems (ADAS) and automated driving systems. It allows users to test core functionalities such as perception, path planning, and vehicle control. One of its key features is that HERE HD live map data \cite{HERE:00} and OpenDRIVE® road networks \cite{opendrive:00} can be imported into MATLAB and these can be used for various design and testing purposes. Further, the users can build photo-realistic 3D scenarios and model various sensors. It is also equipped with a built-in visualizer that allows to view live sensor detection and tracks \cite{Matlab}.

In addition to serving as a simulation and design environment, it also enables users to automate the labeling of objects through the Ground Truth Labeler app \cite{matlab3}. This data can be further used for training purposes or to evaluate sensor performance. 

MATLAB provides several examples on how to simulate various ADAS features including Adaptive Cruise Control (ACC), Automatic Energency Braking (AEB), Automatic Parking Asisst, etc \cite{matlab2}. Last but not the least, the toolbox supports Hardware In the Loop (HIL) testing and C/C++ code generation, which enables faster prototyping. 

\subsection{CarSim}
CarSim is a vehicle simulator commonly used by industry and academia.  The newest version of CarSim supports moving objects and sensors that benefit simulations involving ADAS   and Autonomous Vehicles (AVs)\cite{Carsim}.  In terms of traffic and target objects,  in addition to simulated vehicle, there are up to 200 objects with independent locations and motions. These objects include  static objects such as trees and buildings and dynamic objects such as pedestrians, vehicles, animals, and other objects of interest for ADAS scenarios.

The dynamic object is defined based on the location and orientation that is important in vehicle simulation. Additionally, when the sensors are combined with objects, the objects are considered as targets that can be detected. 

The moving objects can be linked to 3D objects with their own embedded animations, such as walking pedestrians or pedaling bicyclists. If there are ADAS sensors in the simulation, each object has a shape that influences the detection. The shapes may be rectangular, circular, a straight segment (with limited visibility, used for signs), or polygonal. 

In terms of vehicle control, there are several math models available to use and the users can control the motion  using built-in options, either with CarSim commands, or with external models (e.g., Simulink).

The key feature of CarSim is that it has interfaces to other software such as Matlab and LabVIEW. CarSim offers several examples of simulations and it has a detailed documentation, however, it is not an open source simulator.

\subsection{PreScan}
PreScan provides a simulation framework to design ADAS and autonomous driving vehicles. It enables manufacturers to test their intelligent systems by providing a variety of virtual traffic conditions and realistic environments. This is benefited by PreScan's automatic traffic generator. Moreover, PreScan enables users to build their customized sensor suites, control logic and collision warning features. 

PreScan also supports Hardware In the Loop (HIL) simulation which is a common practice for evaluating Electronic Control Unit (ECU). PreScan is good at physics-based calculations of the sensor inputs. The sensor signals are input to the
ECU to evaluate various algorithms. Further, the signals can also be output to either the loop or camera HIL for driver. It also supports real-time data and GPS vehicle data recording, which can then be replayed later on. This is very helpful for situations which are otherwise not easy to simulate with synthetic data. 

Additionally, PreScan offers a unique feature called Vehicle Hardware-In-the-Loop (VeHIL) laboratory. It allows users to create a combined real and virtual system where the test/ego vehicle is placed on a roller bench and other vehicles are represented by wheeled robots with a car-like appearance. The test vehicle is equipped with realistic sensors. By using this combination of ego vehicle and mobile robots, VeHIL is capable of providing detailed simulations for ADAS.

\subsection{CARLA}
CARLA \cite{dosovitskiy2017carla} is an open-source simulator that democratizes autonomous driving research area. The simulator is open source and is developed based on the Unreal Engine\cite{unreal}. It serves as a modular and flexible tool equipped with a powerful API to support training, and validation of ADAS systems. Therefore, CARLA tries to meet the requirements of various use cases of ADAS, for instance training the perception algorithms or learning driving policies. CARLA is developed from scratch  based on the Unreal Engine to execute the simulation and it leverages the OpenDRIVE standard to define roads and urban settings. The CARLA API is customizable by users and provides control over the simulation. It is based on Python and C++ and is constantly growing concurrently with the project that is an ecosystem of projects, built around the main platform by the community.

The CARLA consists of a scalable client-server architecture. The simulation related tasks are deployed at the server including the updates on the world-state and its actors, sensor rendering ,and computation of physics etc.  In order to generate realistic results, the server should run with the dedicated GPU. The client side consists of some client modules that control the logic of agents appearing in the scene including the pedestrians, vehicles, bicycles, motorcycles. Also, the client modules are responsible for the world conditions setups. The setup of all client modules is achieved by using the CARLA API. The 
vehicles, buildings and urban layouts are couple of open digital assets that CARLA provides. In addition, the environmental conditions such as different weather conditions and flexible specification of sensor suits are supported. In order to accelerate queries (such as the closest waypoint in a road), CARLA makes use of RTrees. 

In recent versions, CARLA has more accurate vehicle volumes and more realistic core physics (such as wheel’s friction, suspension, and center of mass). This is helpful when a vehicle turns or a collision occurs. In addition, the process of adding traffic lights and stop signs to the scene have been changed from manual to automatic by leveraging the information provided by the OpenDRIVE file.   

CARLA proposes a safety assurance module based on the RSS library. The responsibility of this module is to put holds on the vehicle controls based on the sensor information. In other words, the RSS defines various situations based on sensor data and then determines a proper response according to safety checks. A situation describes the state of the ego vehicle with an element of the environment. Leveraging the OpenDrive signals enables the RSS module to take different road segments into consideration that helps to check the priority and safety at junctions.

\subsection{Gazebo}
Gazebo is an open source, scalable, flexible and multi-robot 3D simulator \cite{koenig2004design}. It is supported on multiple operating systems, including Linux and Windows. It supports the recreation of both indoor and outdoor 3D environments. 

Gazebo relies on three main libraries, which include, physics, rendering, and a communication library. Firstly, the physics library allows the simulated objects to behave as realistically as possible to their real counterparts by letting the user define their physical properties such as mass, friction coefficient, velocity, inertia, etc. Gazebo uses Open Dynamic Engine (ODE) as its default physics engine but it also supports others such as Bullet, Simbody and Dynamic Open Source Physics Engine (DART). Secondly, for visualization, it uses a rendering library called Object-Oriented Graphics Rendering Engine (OGRE), which makes it possible to visualize dynamic 3D objects and scenes. Thirdly, the communication library enables communication amongst various elements of Gazebo. Besides these three core libraries, Gazebo offers plugin support that allows the users to communicate with these libraries directly.

There are two core elements that define any 3D scene. In Gazebo terminology, these are called a world and a model. A world is used to represent a 3D scene, which could be an indoor or an outdoor environment. It is a user defined file in a Simulation Description File (SDF) format \cite{sdf}, with a dot world extension. The world file consists of one or many models. Further, a model is any 3D object. It could be a static object such as a table, house, sensor, or a robot or a dynamic object. The users are free to create objects from scratch by defining their visual, inertial and collision properties in an SDF format. Optionally, they can define plugins to control various aspects of simulation, such as, a world plugin controls the world properties and model plugin controls the model properties and so on. It is important to note that Gazebo has a wide community support that makes it possible to share and use models already created by somebody else. Additionally, it has well-maintained documentation and numerous tutorials.

Finally, Gazebo is a standalone simulator. However, it is typically used in conjunction with ROS \cite{takaya2016simulation}, \cite{yao2015simulation}. Gazebo supports modelling of almost all kinds of robots. \cite{cardemo} presents a complex scenario that shows the advanced modeling capabilities of Gazebo which models Prius Hybrid model of a car driving in the simulated M-city. 

\subsection{LGSVL}
LG Electronics America R\&D Center (LGSVL) \cite{rong2020lgsvl} is a multi-robot AV simulator. It proposes an out-of-the-box solution for the AV algorithms to test the autonomous vehicle algorithms. It is integrated to some of the platforms that make it easy to test and validate the entire system. The simulator is open source and is developed based on the Unity game engine \cite{unity}. LGSVL provides different bridges for message passing between the AD stack and the simulator backbone. 

The simulator has different components. The user AD stack that provides the development, test, and verification platform to the AV developers. The simulator supports ROS1, ROS2, and Cyber RT messages. This helps to connect the simulator to the Autoware\cite{Autoware} and Baidu Apollo \cite{Apollo} which are the most  popular AD stacks. In addition, multiple AD simulators can communicate simultaneously with the simulator via ROS and ROS2 bridges for the Autoware and customized bridge for Baidu Apollo. LGSVL Simulator leverages Unity’s game engine that helps to generate photo-realistic virtual environments based on High Definition Render Pipeline (HDRP) technology. The simulation engine provides different functions for simulating the  environment simulation (traffic simulation and physical environment simulation), sensor simulation, and vehicle dynamics. The simulator provides a Python API to control different environment entities. In addition, sensor and vehicle models proposes a customizable set of sensors via setting up a JSON file that enables the specification of the intrinsic and extrinsic parameters. The simulator currently supports camera, LiDAR, IMU, GPS, and radar. additionally, developers can define customized sensors. The simulator provides various options, for instance the segmentation and semantic segmentation. Also, LGSVL provides Functional Mockup Interface (FMI) in order to integrate vehicle dynamics model platform to the external third party dynamics models. 
Finally, the weather condition, the day time, traffic agents, and dynamic actors are specified based on 3D environment. One of the important features of LGSVL is exporting HD maps from 3D environments.
\section{Comparison}
\label{sec:evaluation}
This section provides a comparison of different simulators described under Section 4, starting with MATLAB, CarSim and PreScan. Then we compare Gazebo and CARLA, followed by comparison of CARLA and LGSVL. Finally, we conclude with our analysis and key observations.

MATLAB/Simulink is designed for the simple scenarios. It is good at computation and has efficient plot functions. The capability of co-simulation with other software like CarSim makes it  easier to build various vehicle models. It is common to see users using the vehicle models from CarSim and build their upper control algorithms in MATLAB/Simulink to do a co-simulation project. However, MATLAB/Simulink has limited ability to realistically visualize the traffic scenarios, obstacles and pedestrian models. PreScan has strong capability to simulate the environment of the real world such as the weather conditions that MATLAB/Simulink and CarSim cannot do, it also has interfaces with MATLAB/Simulink that makes modelling more efficient.

Further, Gazebo is known for its high flexibility and its seamless integration with ROS. While the high flexibility is advantageous because it gives full control over simulation, it comes at a cost of time and effort. As opposed to CARLA and LGSVL simulators, the creation of a simulation world in Gazebo is a manual process where the user must create 3D models and carefully define their physics and their position in the simulation world within the XML file. Gazebo does include various sensor models and it allows users to create new sensor models via the plugins. Next, we compare CARLA and LGSVL simulators. 

Both CARLA and LGSVL provide high quality simulation environments that require GPU computing unit in order to run with reasonable performance and frame rate.  The user can invoke different facilities in CARLA and LGSVL via using a flexible API. Although, the facilities are different between two simulators. For instance, CARLA provides build-in recorder while LGSVL does not provide. Therefore, in order to record videos in LGSVL, the user can leverage video recording feature in Nvidia drivers. CARLA and LGSVL provide a variety of sensors, some of these sensors are common between them such as  Depth camera, Lidar, and IMU. In addition, each simulator provides different sensors with description provided in their official website. Both simulators, CARLA and LGSVL, enhance users to create the custom sensors. The new map generation has different process in CARLA and LGSVL. The backbone of CARLA simulator is Unreal Engine that generates new maps by automatically adding stop signs based on the OpenDRIVE technology. On the other hand, the backbone of LGSVL simulator is Unity game engine and user can generate new map by manually importing different components into the Unity game engine. Additionally, the software architecture in CARLA and LGSVL is quite different. LGSVL mostly connects to AD stacks (Autoware, Apollo, and etc. ) based on different bridges and most of the simulators' facilities publishes or subscribes the data on specified topics in order to enables AD stacks to consume data. On the other hand, most of facilities in CARLA are built in, although it enables users to connects to ROS1,ROS2, and Autoware via using the bridges.

While all of the six simulators described in the paper offer their own advantages and disadvantages, we make the following key observations. 
\begin{itemize}
  \item \textbf{\emph{Observation 1:}} LGSVL and CARLA are most suited for end to end testing of unique functionalities that self-driving cars offer such as perception, mapping, localization, and vehicle control because of many built-in automated features they support.
  \item \textbf{\emph{Observation 2:}} Gazebo is a popular robotic simulator but the time and effort needed to create dynamic scenes does not make it the first choice for testing end to end systems for self-driving cars
  \item \textbf{\emph{Observation 3:}} MATLAB/Simulink is one of the best choices for testing upper level algorithms because of the clearly presented logic blocks in Simulink. Additionally, it has a fast plot function that makes it easier to do the results analysis.
  \item \textbf{\emph{Observation 4:}} CarSim specializes in vehicle dynamic simulations because of its complete vehicle library and variety of vehicle parameters available to tune. However, it has limited ability to build customized upper-level algorithms in an efficient way.
  \item \textbf{\emph{Observation 5:}} PreScan has a strong capability of building realistic environments and simulating different weather conditions.
\end{itemize}

In Table~\ref{tab:comparison}, we provide a comparison summary where all six simulators described in this paper are further compared.

\begin{table*}[h!]
\caption{Comparison of various simulators.}
\label{tab:comparison}
\centering
\begin{tabular}{|p{0.15\linewidth}||p{0.1\linewidth}|p{0.1\linewidth}|p{0.1\linewidth}|p{0.1\linewidth}|p{0.1\linewidth}|p{0.1\linewidth}|}
\hline
\textbf{Requirements} & \textbf{MATLAB (Simulink)} & \textbf{CarSim} & \textbf{PreScan} & \textbf{CARLA} & \textbf{Gazebo} & \textbf{LGSVL}\\
\hline
\hline
\textbf{Perception}: Sensor models supported & Y & Y  & Y & Y(1) & Y(2) & Y(3) \\
\hline
\textbf{Perception}: support for different weather conditions & N & N  & Y  & Y & N & Y \\
\hline
\textbf{Camera Calibration} & Y & N  & Y & Y & N  & N \\
\hline
\textbf{Path Planning} & Y & Y  & Y  & Y & Y & Y \\
\hline
\textbf{Vehicle Control}: Support for proper vehicle dynamics & Y & Y  & Y  & Y & Y & Y(3) \\
\hline

\textbf{3D Virtual Environment} & U & Y  & Y  & Y, Outdoor (Urban) & Y, Indoor and Outdoor & Y, Outdoor (Urban)\\
\hline
\textbf{Traffic Infrastructure} & Y, allows to build lights model & Y  & Y & Y, Traffic lights, Intersections, Stop signs, lanes  & Y, allows to manually build all kinds of models & Y\\
\hline
\textbf{Traffic Scenario simulation}: Support of different types of Dynamic objects & Y & Y & Y & Y & N(2) & Y \\
\hline
\textbf{2D/3D Ground Truth} & Y & N & N & Y & U & Y\\
\hline
\textbf{Interfaces to other software} & Y,  with Carsim, Prescan,ROS & Y, with Matlab(Simulink) & Y, with MATLAB(Simulink) & Y, with ROS, Autoware & Y, with ROS & Y, with Autoware, Apollo, ROS\\
\hline
\textbf{Scalability} via a server multi-client architecture & U & U & U & Y & Y & Y \\
\hline 
\textbf{Open Source} & N & N & N & Y & Y & Y \\
\hline
\textbf{Well-maintained/Stable} & Y & Y & Y & Y & Y & Y \\
\hline
\textbf{Portability} & Y & Y & Y & Y, Windows, Linux & Y, Windows, Linux & Y, Windows, Linux\\
\hline
\textbf{Flexible API} & Y & Y & U & Y (2) & Y & Y\\
\hline
\end{tabular}
\begin{tablenotes}
    \item[1] Y=supported, N = not supported, U = unknown
    \item[2] (1) See section 4.4 for details about Carla.
    \item[3] (2) See section 4.5 for details about Gazebo
    \item[4] (3) See section 4.6 for details about LGSVL
\end{tablenotes}
\end{table*}

\section{Challenges}
The automotive simulators have come a long way. Although simulation has now become a cornerstone in the development of self-driving cars, common standards to evaluate simulation results is lacking. For example, the Annual Mileage Report submitted to the California Department of Motor Vehicle by the key players such as Waymo, Cruise, and Tesla does not include the sophistication and diversity of the miles collected through simulation \cite{amr}. It would be more beneficial to have simulation standards that could help make a more informative comparison between various research efforts. 

Further, we are not aware of any simulators that are currently capable of testing the concept of connected vehicles, where vehicles communicate with each other and with the infrastructure. However, there are test beds available such as the ones mentioned in the report \cite{dotintelligent} from the US Department of Transportation.

In addition, current simulators, for instance CARLA and LGSVL, are on-going projects and add the most recent technologies. Therefore, the user may encounter with undocumented errors or bugs. Therefore, the community support is quite important which can improve the quality of open source simulators and ADAS tests.
\section{Related Work}
\label{sec:related}
There are many other simulators available that are not explicitly reviewed in this paper. For example, RoadView is a traffic scene modelling simulator built using image sequences and the road Global Information System (GIS) data \cite{zhang2014roadview}. \cite{banerjee2019development} provides an in-depth review of CARLA simulator and how it can be used to test autonomous driving algorithms. Similarly, \cite{fadaie2019state}, \cite{chao2020survey}, and \cite{figueiredo2009approach} provide review of various other automotive and robotic simulators. \cite{tang2017distributed} discusses a distributed simulation platform for testing.

\section{Conclusion Remarks}
\label{sec:conclusion}
In this paper, we compare MATLAB/Simulink, CarSim, PreScan, Gazebo, CARLA and LGSVL simulators for testing self-driving cars. The focus is on how well they are at simulating and testing perception, mapping and localization, path planning and vehicle control for the self-driving cars. Our analysis yields five key observations that are discussed in Section 5. We also identify key requirements that state-of-the-art simulators must have to yield reliable results. Finally, several challenges still remain with the simulation strategies such as the lack of common standards as mentioned in Section 6. In conclusion, simulation will continue to help design self-driving vehicles in a safe, cost effective, and timely fashion, provided the simulations represent the reality.

\newpage
\bibliographystyle{plain}
\bibliography{reference/mist}
\end{document}